\documentclass[runningheads]{llncs}
\usepackage{graphicx}
\usepackage{amsmath,amssymb} 
\usepackage{color}
\usepackage[table]{xcolor}
\usepackage{tabularx}
\usepackage{booktabs}
\usepackage{pdfcomment}
\usepackage{color}
\usepackage{bm}
\definecolor{lightbluishgrey}{rgb}{0.9,0.91,0.95}

\hypersetup{citecolor=black}
\newcommand{\cheatvspace}[1]{}

\usepackage{wrapfig}

\newcommand{\refequ}[1] {Equation~(\ref{equ:#1})}
\newcommand{\reffig}[1] {Figure~\ref{fig:#1}}

\newcommand{\refsec}[1] {Section~\ref{sec:#1}}

\let\mat = \mathbf
\usepackage{xfrac}

\usepackage{amsmath}
\usepackage{amssymb}    
\usepackage{cancel}
\usepackage[T1]{fontenc}

\newcommand{\R}{\mathbb{R}}
\newcommand{\vc}[1]{\mathbf{#1}}

\renewcommand{\b}{\vc{b}}

\renewcommand{\d}{\vc{d}}

\newcommand{\p}{\vc{p}}

\renewcommand{\t}{\vc{t}}
\renewcommand{\u}{\vc{u}}

\newcommand{\x}{\vc{x}}

\newcommand{\B}{\mat{B}}
\newcommand{\D}{\mat{D}}

\newcommand{\I}{\mat{I}}

\newcommand{\M}{\mat{M}}

\renewcommand{\P}{\mat{P}}

\renewcommand{\S}{\mat{S}}

\newcommand{\argmin}{\mathop{\text{argmin }}}

\usepackage{siunitx}
\usepackage[mathletters]{ucs}
\usepackage[utf8x]{inputenc}
\usepackage{upgreek}
\usepackage{dblfloatfix}

\usepackage{graphicx}

\usepackage{wrapfig}

\makeatletter
\let\save@mathaccent\mathaccent
\newcommand*\if@single[3]{%
  \setbox0\hbox{${\mathaccent"0362{#1}}^H$}%
  \setbox2\hbox{${\mathaccent"0362{\kern0pt#1}}^H$}%
  \ifdim\ht0=\ht2 #3\else #2\fi
  }
\newcommand*\rel@kern[1]{\kern#1\dimexpr\macc@kerna}
\newcommand*\widebar[1]{\@ifnextchar^{{\wide@bar{#1}{0}}}{\wide@bar{#1}{1}}}
\newcommand*\wide@bar[2]{\if@single{#1}{\wide@bar@{#1}{#2}{1}}{\wide@bar@{#1}{#2}{2}}}
\newcommand*\wide@bar@[3]{%
  \begingroup
  \def\mathaccent##1##2{%
    \let\mathaccent\save@mathaccent
    \if#32 \let\macc@nucleus\first@char \fi
    \setbox\z@\hbox{$\macc@style{\macc@nucleus}_{}$}%
    \setbox\tw@\hbox{$\macc@style{\macc@nucleus}{}_{}$}%
    \dimen@\wd\tw@
    \advance\dimen@-\wd\z@
    \divide\dimen@ 3
    \@tempdima\wd\tw@
    \advance\@tempdima-\scriptspace
    \divide\@tempdima 10
    \advance\dimen@-\@tempdima
    \ifdim\dimen@>\z@ \dimen@0pt\fi
    \rel@kern{0.6}\kern-\dimen@
    \if#31
      \overline{\rel@kern{-0.6}\kern\dimen@\macc@nucleus\rel@kern{0.4}\kern\dimen@}%
      \advance\dimen@0.4\dimexpr\macc@kerna
      \let\final@kern#2%
      \ifdim\dimen@<\z@ \let\final@kern1\fi
      \if\final@kern1 \kern-\dimen@\fi
    \else
      \overline{\rel@kern{-0.6}\kern\dimen@#1}%
    \fi
  }%
  \macc@depth\@ne
  \let\math@bgroup\@empty \let\math@egroup\macc@set@skewchar
  \mathsurround\z@ \frozen@everymath{\mathgroup\macc@group\relax}%
  \macc@set@skewchar\relax
  \let\mathaccentV\macc@nested@a
  \if#31
    \macc@nested@a\relax111{#1}%
  \else
    \def\gobble@till@marker##1\endmarker{}%
    \futurelet\first@char\gobble@till@marker#1\endmarker
    \ifcat\noexpand\first@char A\else
      \def\first@char{}%
    \fi
    \macc@nested@a\relax111{\first@char}%
  \fi
  \endgroup
}
\makeatother

\usepackage[width=122mm,left=12mm,paperwidth=146mm,height=193mm,top=12mm,paperheight=217mm]{geometry}
\usepackage{caption}
\usepackage{hyperref}
\hypersetup{
     colorlinks   = true,
     citecolor    = black,
     linkcolor = black,
    urlcolor=black
}

\begin{document}
\pagestyle{headings}
\def\ECCV16SubNumber{1622}  

\title{Real-Time Facial Segmentation\\ and Performance Capture from RGB Input} 

\titlerunning{Real-Time Facial Segmentation and Performance Capture from RGB Input}

\authorrunning{Shunsuke Saito, Tianye Li, Hao Li}

\author{Shunsuke Saito \qquad Tianye Li \qquad Hao Li}
\institute{Pinscreen \qquad University of Southern California}

\maketitle

\begin{abstract}
We introduce the concept of unconstrained real-time 3D facial performance capture through explicit semantic segmentation in the RGB input. To ensure robustness, cutting edge supervised learning approaches rely on large training datasets of face images captured in the wild. While impressive tracking quality has been demonstrated for faces that are largely visible, any occlusion due to hair, accessories, or hand-to-face gestures would result in significant visual artifacts and loss of tracking accuracy. The modeling of occlusions has been mostly avoided due to its immense space of appearance variability. To address this curse of high dimensionality, we perform tracking in unconstrained images assuming non-face regions can be fully masked out. Along with recent breakthroughs in deep learning, we demonstrate that pixel-level facial segmentation is possible in real-time by repurposing convolutional neural networks designed originally for general semantic segmentation. We develop an efficient architecture based on a two-stream deconvolution network with complementary characteristics, and introduce carefully designed training samples and data augmentation strategies for improved segmentation accuracy and robustness. We adopt a state-of-the-art regression-based facial tracking framework with segmented face images as training, and demonstrate accurate and uninterrupted facial performance capture in the presence of extreme occlusion and even side views. Furthermore, the resulting segmentation can be directly used to composite partial 3D face models on the input images and enable seamless facial manipulation tasks, such as virtual make-up or face replacement.
\end{abstract}

\section{Introduction}\label{sec:intro}

Recent advances in real-time 3D facial performance capture~\cite{weise2011realtime,Bouaziz:2013:OMR,li2013realtime,Cao:2013:SRR,cao2014displaced,cao2015real,hsieh2015unconstrained} have not only transformed the entertainment industry with highly scalable animation and affordable production tools~\cite{Faceshift:2014:FS}, but also popularized mobile and social media apps with facial manipulation and analytics software. 
The key factors behind this democratization are: (1) the ability to capture compelling facial animations in real-time and (2) the accessibility of commodity sensors (video and RGB-D). Many state-of-the-art techniques have been developed to operate robustly in natural environments, but pure RGB solutions are still susceptible to occlusions (e.g., caused by hair, hand-to-face gestures, or accessories), which result in unpleasant visual artifacts or the inability to correctly initialize facial tracking. 

While it is known that the shape and appearance of fully visible faces can be represented compactly through linear models~\cite{Blanz:1999:MMS,cao2014facewarehouse}, any occlusion or uncontrolled illumination could cause high non-linearities to a 3D face fitting problem. As this space of variation becomes intractable, supervised learning methods have been introduced to predict facial shapes through large training datasets of face images captured under unconstrained and noisy conditions. We observe that if such \emph{occlusion noise} can be fully eliminated, the dimensionality of facial modeling could be drastically reduced to that of a well-posed and constrained problem. In other words, if reliable dense facial segmentation  is possible, 3D facial tracking from RGB input becomes a significantly easier problem. Only recently has the deep learning community demonstrated highly effective semantic segmentations, such as the fully convolutional network (FCN) of~\cite{long_shelhamer_fcn} or the deconvolutional network (DeconvNet) of~\cite{noh2015learning}, by repurposing highly efficient classification networks~\cite{Chatfield14,Krizhevsky_imagenetclassification} for dense predictions of general objects (e.g., humans, cars, etc.). 

We present a real-time facial performance capture approach by explicitly segmenting facial regions and processing masked RGB data. 
We rely on the effectiveness of deep learning to achieve clean facial segmentations in order to enable robust facial tracking under severe occlusions.
We propose an end-to-end segmentation network that also uses a two-stream deconvolution network with complementary characteristics, but shares the lower convolution network to enable real-time performance. A final convolutional layer recombines both outputs into a single probability map which is converted into a refined segmentation mask via graph cut algorithm~\cite{Rother:2004:GIF}.
Our 3D facial tracker is based on a state-of-the-art displaced dynamic expression (DDE) method~\cite{cao2014displaced} trained with segmented input data.
%
%
While the network parameters of pre-learned representations are transferred to our facial segmentation task, we fine tune the network using training samples from the LFW~\cite{LFWTech} and FaceWarehouse~\cite{cao2014facewarehouse} datasets. Furthermore, separating facial regions from occluding objects with similar colors and fine structures (e.g. hands) is still extremely challenging for any existing segmentation network, since no such supervision is provided by existing data sets. We propose a data augmentation strategy based on perturbations, croppings, occlusion generation, hand compositings, as well as the use of negative samples containing no faces. Once our dense prediction model is trained, we replace the training database for DDE regression with masked faces obtained from our convolutional network.

Our approach retains every capability of Cao et al.~\cite{cao2014displaced}'s algorithm such as real-time performance and the absence of a calibration process, but considerably enhances its robustness w.r.t. occlusions and even side views. We demonstrate uninterrupted tracking in the presence of highly challenging occlusions such as hands which have similar skin tones as the face and fine scale boundary details. Furthermore, our facial segmentation solution provides masked images which enables interesting compositing effects such as tracked facial models under hair and other occluding objects. These capabilities were only demonstrated recently using a robust geometric model fitting approach on depth sensor data~\cite{hsieh2015unconstrained}. Since we only assume RGB data as input, our method not only addresses a fundamental challenge of real-time facial segmentation, but also provides unmatched flexibility for deployment, and requires minimal implementation effort.

\vspace{0.5cm}

We make the following contributions:
\begin{itemize}
	\item We present the first real-time facial segmentation framework from pure RGB input using a convolutional neural network. We demonstrate the importance of carefully designed datasets and data augmentation strategies for handling challenging occlusions such as hands.
	\item We improve the efficiency and accuracy of existing segmentation networks using an architecture based on two-stream deconvolution networks and shared convolution network.
	\item We demonstrate superior tracking accuracy and robustness through explicitly facial segmentation and regression with masked training data, and outperform the current state-of-the-art.
\end{itemize}

\section{Related Work}\label{sec:related}

The fields of facial tracking and animation have undergone a long thread of major research milestones
in both, the vision and graphics community, as well as influencing the industry widely over the past two decades.

In high-end film and game production, performance-driven techniques are commonly 
used to scale the production of realistic facial animation. An overview is discussed in 
Pighin and Lewis~\cite{PighinLewisCourse06}. 
To meet the high quality bars, techniques for production
typically build on sophisticated sensor equipments and controlled capture settings~\cite{Guenter:1998:MF,zhang-siggraph2004-stfaces,FurukawaP09,Li:2009:RSV,Bee11,fyffe2011comprehensive,Bhat:2013:HFF,Fyffe:2014:DHF}. While exceptional tracking accuracy can be achieved, these methods are generally computationally expensive and the full visibility of the face needs to be ensured.

On the other extreme, 2D facial tracking methods that work in fully unconstrained settings have been explored extensively for applications such as face recognition and emotion analytics. Even though only sparse 2D facial landmarks are detected, many techniques are designed to be robust to uncontrolled poses, challenging lighting conditions, and rely on a single-view 2D input. Early algorithms are based on parametric models~\cite{li93threed,bregler1994surface,Black:1995:TRR,Essa:1996:MTI,Decarlo:2000:OFC}, but later outperformed by more robust and real-time data-driven methods such as active appearance models (AAM)~\cite{cootes2001active} and constrained local models (CLM)~\cite{Cristinacce:2008:AFL}. While the landmark mean-shift approach of~\cite{Saragih:2011:DMF} and the supervised descent method of~\cite{xiong2013supervised} avoid the need of user-specific training, more efficient solutions exist based on explicit shape regressions~\cite{export:206636,kazemi2014one,ren2014face}. However, these methods are all sensitive to occlusions and only a limited number of 2D features can be detected. 

Weise and colleagues~\cite{weise2009face} demonstrated the first system to produce compelling facial performance capture in real-time using a custom 3D depth sensor based on structured light. The intensive training procedure was later reduced significantly using an example-based algorithm developed by Li and collaborators~\cite{li2010example}. With consumer depth sensors becoming mainstream (e.g., Kinect, Realsense, etc.), a whole line of real-time facial animation research have been developed with focus on deployability. The work of~\cite{weise2011realtime} incorporated pre-recorded motion priors to ensure stable tracking for noisy depth maps, which resulted in the popular animation software, Faceshift~\cite{Faceshift:2014:FS}. By optimizing the identity and expression models online, Li and coworkers~\cite{li2013realtime}, as well as Bouaziz and collaborators~\cite{Bouaziz:2013:OMR} eliminated the need of user-specific calibration. For uninterrupted tracking under severe occlusions, Hsieh and colleagues~\cite{hsieh2015unconstrained} recently proposed an explicit facial segmentation technique based on depth and RGB cues. While the idea of explicitly segmenting faces is similar to our work, their method relies on depth sensor input.

While the generation of 3D facial animations from pure RGB input have been demonstrated using sparse 2D landmarks detection~\cite{Pighin:etal:1999,Chuang02Performance,Chai:2003:VCF}, a superior performance capture fidelity and robustness has only been shown recently by Cao and coworkers~\cite{Cao:2013:SRR} using a 3D shape regression approach. Cao and colleagues~\cite{cao2014displaced} later extended the efficient two-level boosted regression technique introduced in~\cite{export:206636} to the 3D case in order to avoid user-specific calibration. Higher fidelity facial tracking from monocular video has also been demonstrated with additional high-resolution training data~\cite{cao2015real}, very large datasets of a person~\cite{totalmoving}, or more expensive non-real-time computation~\cite{garrido2013reconstructing,shi2014automatic}. While robust to unconstrained lighting environments and large head poses, these methods are sensitive to large occlusions and cannot segment facial regions.

Due to the immense variation of facial appearances in unconstrained images, it is extremely challenging to obtain clean facial segmentations at the pixel level.  
The hierarchical CNN-based parsing network of Luo and collaborators~\cite{luo2012hierarchical} generates masks of individual facial components such as eyes, nose, and mouth even in the presence of occlusions, but does not segment the facial region as a whole. Smith and coworkers~\cite{smith2013exemplar} use an example-based approach for facial region and component segmentation, but the method requires sufficient visibility of the face. These two methods are computationally intensive and susceptible to wrong segmentations when occlusions have similar colors as the face. By alternating between face mask prediction and landmark localization with deformable part models, Ghiasi and Fowlkes~\cite{ghiasi2015using} have recently demonstrated state-of-the-art facial segmentation results on the Caltech Occluded Faces in the Wild (COFW) dataset~\cite{burgos2013robust} at the cost of expensive computations. 
Without explicitly segmenting the face, occlusion handling methods have been proposed for the detection of 2D landmarks within an AAM frameworks~\cite{Gross:2006:AAM}, but superior results were later shown using techniques based on discriminatively trained deformable parts model ~\cite{Ramanan:2012:FDP,ghiasi2014occlusion}. Highly efficient landmark detection has been recently demonstrated using cascade of regressors trained with occlusion data~\cite{burgos2013robust,YuLBM14}. 

\section{Overview}\label{sec:system}

\begin{wrapfigure}{r}{0.45\textwidth}
\vspace{-0.8cm}
\centering
\includegraphics[width=0.43\textwidth]{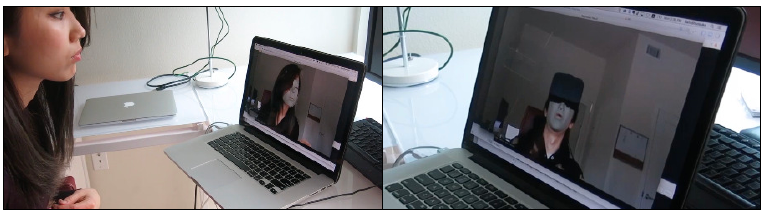}
\caption{Capture setting using a laptop integrated RGB webcam.\label{fig:system}}
\vspace{-0.5cm}
\end{wrapfigure}

\begin{figure*}[t!]
 \centering
\includegraphics[width=\textwidth]{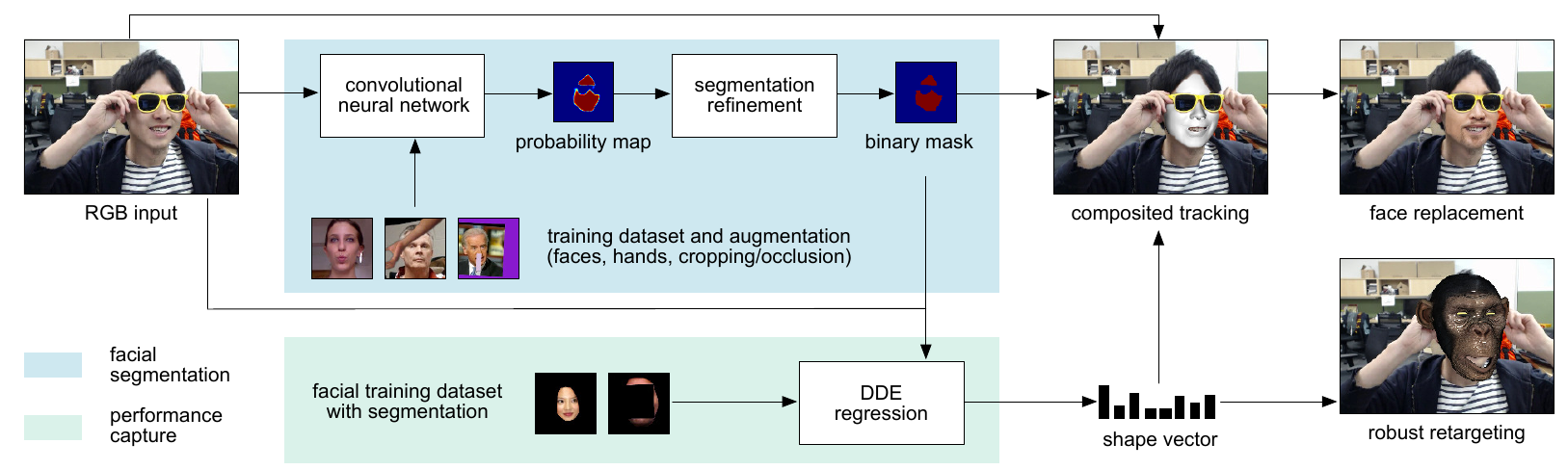}
\caption{Overview of our facial segmentation and performance capture pipeline. 
 \label{fig:overview}}
\end{figure*}


As illustrated in~\reffig{overview}, our system is divided into a facial segmentation stage (blue) and a performance capture stage (green).
Our pipeline takes an RGB image as input and produces a binary segmentation mask in addition to a tracked 3D face model, which is parameterized by a shape vector, as output.
The binary mask represents a per-pixel facial region estimated by a deep learning framework for facial segmentation. 
Following Cao et al.'s DDE regression technique~\cite{cao2014displaced}, the shape vector describes the rigid head motion and the facial expression coefficients, which drive the 
animation of a personalized 3D tracking model. In addition, the shape of the user's identity and the focal length are solved concurrently during performance capture.
While the resulting tracking model represents the 
shape of the subject, the shape vector can be used to retarget any digital character with compatible animation controls as input.

Our convolutional neural network first predicts a probability map on a cropped rectangular face region for which size and positions are determined based on the bounding box of the projected 3D tracking model from the previous frame. The face region of the initial frame is detected using the method of Viola and Jones~\cite{violajones}. The output probability map is a smaller fixed-size resolution 
image ($128\times 128$ pixels) and describes the likelihood for each pixel being labeled as part of the specific face region. While two output maps (one for the overall shape and one for fine-scaled details) are simultaneously produced by our two-stream deconvolution network, a single output probability map is generated through a final convolutional layer.
To ensure accurate and robust facial segmentation, we train our convolutional neural network using a large dataset of segmented face images, augmented with peturbations, synthetic occlusions, croppings, and hand compositings, as well as negative samples containing no faces. We convert the resulting probability map into a binary mask using a graph cut algorithm~\cite{gridcut} and bilinearly upsample the mask to the original input resolution. 

We then use this segmentation mask as input to the facial tracker as well as for compositing partial 3D facial models during occlusions. This facial segmentation technique is also used to produce training data for the regression model of the DDE framework. Our facial performance capture pipeline is based on the state-of-the-art method of~\cite{cao2014displaced}, which does not require any calibration step for individual users. The training process and the regression explicitly take the segmentation mask into account. Our system runs in real-time on commercially available desktop machines with sufficiently powerful GPU processors. For many mobile devices such as laptops, which are not yet ready for deep neural net computations, we can optionally 
offload the segmentation processing over Wi-Fi to a desktop machine with high-end GPU resources for real-time performance.

\section{Facial Segmentation}\label{sec:probmap}

Our facial segmentation pipeline computes a binary mask from the bounding box of a face in the input image. The cropped face image is first resized to a small $128\times128$ pixel resolution image, which is passed to a convolutional neural network for a dense 2-class segmentation problem. Similar to state-of-the-art segmentation networks~\cite{long_shelhamer_fcn,noh2015learning,chen2014semantic}, the overall network consists of two parts, (1) a lower convolution network for multi-dimensional feature extraction and (2) a higher deconvolution network for shape generation. This shape corresponds to the segmented object and is reconstructed using the features obtained from the convolution network.
The output is a dense $128\times128$ probability map that assigns each pixel to either a face or non-face region. While both state-of-the-art networks, FCN~\cite{long_shelhamer_fcn} and DeconvNet~\cite{noh2015learning} use the identical convolutional network based on VGG-16 layers~\cite{Simonyan14c}, they approach deconvolution differently. FCN performs a simple deconvolution using a single bilinear interpolation layer, and produces coarse, but clean overall shape segmentations, because the output layer is closely connected to the convolution layers preventing the loss of spatial information. DeconvNet on the other hand, mirrors the convolution process with multiple series of unpooling, deconvolution, and rectification layers, and generates detailed segmentations at the cost of increased noise. Noh and collaborators~\cite{noh2015learning} proposed to combine the outputs of both algorithms through averaging followed by a post-hoc segmentation refinement based on conditional random fields~\cite{NIPS2011_4296}, but the computation is prohibitively intensive. Instead, we develop an efficient network with shared convolution layers to reduce the number of parameters and operations, but split the deconvolution part into a two-stream architecture to benefit from the advantages of both networks. The output probability map resulting from a bilinear interpolation and mirrored deconvolution network are then concatenated before a final convolutional layer merges them into a single high-fidelity output map. We then use a standard graph cut algorithm~\cite{gridcut} to convert the probability map into a clean binary facial mask and upsample to the resolution of the original input image via bilinear interpolation.

\paragraph{Architecture.}

\begin{wrapfigure}{r}{0.45\textwidth}
\vspace{-0.8cm}
\centering
\includegraphics[width=0.43\textwidth]{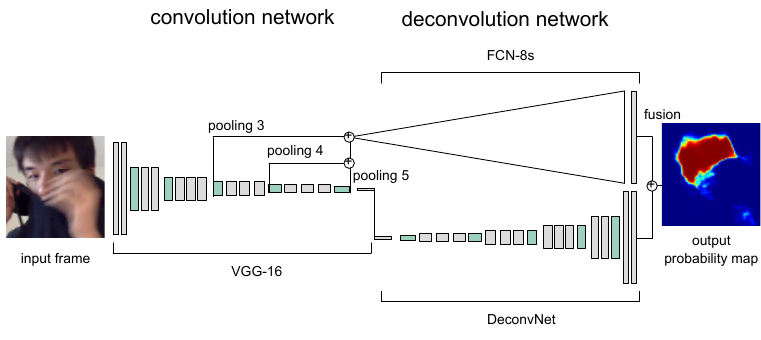}
\caption{ConvNet architecture with two-stream deconvolution network.
 \label{fig:architecture}}
\vspace{-0.5cm}
\end{wrapfigure}

Our segmentation network consists of a single convolution network connected to two different deconvolution networks, DeconvNet and an 8 pixel stride FCN-8s as shown in Figure~\ref{fig:architecture}.
The network is based on a 16 layer VGG architecture and pre-trained on the PASCAL VOC 2012 data set with 20 object categories~\cite{Chatfield14}.
More specifically, VGG has 13 layers of convolutions and rectified linear units (ReLU), 5 max pooling layers, two fully connected layers, and one classification layer.
DeconvNet mirrors the convolutional network to generate a probability map with the same resolution as the input, by applying upsampling operations (deconvolution) and the inverse operation of pooling (unpooling). Even though deconvolution is fast, the runtime performance is blocked by the first fully connected layer which becomes the bottleneck of the segmentation pipeline. To enable real-time performance on a state-of-the-art GPU, we reduce the kernel size of the first fully connected layer from $7\times7$ to $4\times4$ pixels.

Further modifications to the FCN-8s are needed in order to connect the output of both DeconvNet and FCN deconvolution networks to the final convolutional layer. The output size of each deconvolution is controlled by zero padding, so that the size of each upsampled activation layer is aligned with the output of the previous pooling layer. While the original FCN uses the last fully connected layer as the coarsest prediction, we instead use the output of the last pooling layer, as the coarsest prediction in order to preserve spatial information like in DeconvNet. The obtained coarse prediction is then sequentially deconvoluted and fused with the output of pooling layer 4 and 3, and then a deconvolution layer upsamples the fused prediction to the input image size. Since our 2-class labeling problem is considerably less complex than multi-class ones, losing information from discarded layers would not really affect the segmentation accuracy. In the final layer, the output of both deconvolution networks are concatenated into a single matrix and we apply a $1\times1$ convolution to obtain a score map, followed by a softmax operation to produce the final fused probability map. In this way we can even learn blending weights between the two networks as convolution parameters, instead of a simple averaging of output maps as proposed by the separate treatment of~\cite{noh2015learning}. Please refer to the supplemental materials for the detailed configuration of our proposed network.

\paragraph{Training.}

\begin{wrapfigure}{r}{0.45\textwidth}
\vspace{-0.8cm}
\centering
\includegraphics[width=0.43\textwidth]{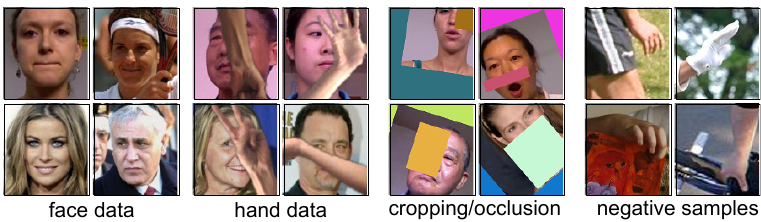}
\caption{Segmentation training data.
 \label{fig:segtraining}}
\vspace{-0.5cm}
\end{wrapfigure}


For an effective facial segmentation in unconstrained images, our convolutional neural network needs to be trained with large image datasets containing face samples and their corresponding ground truth binary masks. The faces should span a sufficiently wide range of shapes, appearance, and illumination conditions. We therefore collect $2927$ images from the LFW face database~\cite{LFWTech} and $5094$ images from the FaceWarehouse dataset~\cite{cao2014facewarehouse}. While the LFW dataset already contains pre-labeled face segmentations, we segment those in FaceWarehouse using a custom semi-automatic tool. We use the available fitted face templates to estimate skin tones and perform a segmentation refinement using a graph cut algorithm~\cite{Rother:2004:GIF}. Each sample is then manually inspected and corrected using additional seeds to ensure that occlusions such as hair and other accessories are properly handled. 

To prevent overfitting, we augment our dataset with additional $82,770$ images using random perturbations of translation, rotation, and scale. 
The data consist of mostly photographs with a large variety of faces in different head poses, expressions, and under different lightings. Occlusions through hair, hands, and other objects are typically avoided. We therefore generate additional $82,770$ samples based on random sized and uniformly colored rectangles on top of each face sample to increase the robustness to partial occlusions (see~\reffig{segtraining}). 

Skin toned objects such as hands and arms are commonly observed during hand-to-face gesticulations but are particularly challenging to segment due to similar colors as the face and fine structures such as fingers. We further augment the training dataset of our convolutional neural network with composited hands on top of the original $8021$ face images. We first captured and manually segmented 1092 hand images of different skin tones, as well as under different lighting conditions and poses. We then synthesized these hand images on top of the original face images, which yields  
$41380$ additional trainining samples using the same perturbation strategy. In total, $132,426$ images were generated to train our network. Our data-augmentation strategy can effectively train the segmentation network and avoid overfitting, even though only limited amount of ground truth data is available.

We initialize the training using pre-trained weights~\cite{Chatfield14} except for the first fully connected layer of the convolution network, since its kernel size is modified for our real-time purposes. Thus, the first fully connected layers and deconvolution layers are initialized with zero-mean Gaussians. The loss function is the sum of softmax functions applied to the output maps of DeconvNet, FCN, and their score maps.
The weights of each softmax function is set to $0.5$, $0.5$, and $1.0$ respectively, and the loss functions are minimized via stochastic gradient descent (SGD) with momentum for stable convergence. 
Notice that by only using the fused score map of DeconvNet and FCN for the loss function, only the DeconvNet model is trained and not FCN. We set $0.01$, $0.9$, and $0.0005$ as the learning rate, momentum, and weight decay, respectively. Our training takes 9 hours using 50,000 SGD iterations on a machine with 16GB RAM and NVIDIA GTX Titan X GPU.

We further fine-tune the trained segmentation by adding negative samples (containing no faces) based on hand, arm, and background images to a random subset of the training data so that the amount of negative samples is equivalent to positive ones. In particular, the public datasets contain images that are both indoor and outdoors. Similar techniques for negative data augmentation has been used previously to improve the accuracy of weak supervision-based classifiers~\cite{siva2012defence,song2014learning}. We use $4699$ hand images that contain no faces from the Oxford hand dataset~\cite{Mittal11}, and further perturb them with random translation and scalings. This fine-tuning with negative samples uses the same loss function and training parameters (momentum, weight decay, and loss weight) as with the training using positive data, but the initial learning rate is changed to 0.001. This training converges after $10,000$ SGD iterations and takes an additional $1.5$ hours of computation.

\paragraph{Segmentation Refinement.}


We convert the  $128\times128$ pixel probability map of the convolutional neural network to a binary mask using a standard graph cut algorithm~\cite{Rother:2004:GIF}.
Even though our facial segmentation is reliable and accurate, a graph cut-based segmentation refinement can purge minor artifacts such as small 'uncertainty' holes at boundaries, which can still appear for challenging cases such as (extreme occlusions, motion blur, etc.).
We optimize the following energy term between adjacent pixels $i$ and $j$ using the efficient GridCut~\cite{gridcut} implementation: 
\begin{equation}\label{equ:graphcut}
{\sum_i{\theta_i}(p_i) - \lambda \sum_{(i, j)} \theta_{i, j} }.
\end{equation}
The unary term $\theta_i(p_i)$ is determined by the facial probability map $p_i$, defined as $\theta_i(p_i) = -\log(p_i)$ for the sink and $\theta_i(p_i) = - \log(1.0 - p_i)$ for the source. The pairwise term $\theta_{i, j} = exp(- \frac{\lVert I_i-I_j \rVert^2}{2\sigma})$, where $I$ is the pixel intensity, $\lambda = 10$, and $\sigma = 5$.
The final binary mask is then bilinearly upsampled to the original cropped image resolution.

\vspace{-0.4cm}
\section{Facial Tracking}\label{sec:tracking}
\vspace{-0.1cm}

After facial segmentation, we capture the facial performance by regressing a 3D face model directly from the incoming RGB input frame. We adopt the state-of-the-art displaced dynamic expression (DDE) framework of~\cite{cao2014displaced} with the two-level boosted regression techniques of~\cite{export:206636} and incorporate our facial segmentation masks into the regression and training process. More concretely, instead of computing the regression on face images with backgrounds and occlusions, where appearance can take huge variations, we only focus on segmented face regions to reduce the dimensionality of the problem. While the original DDE technique is reasonably robust for sufficiently large training datasets, we show that processing accurately segmented images significantly improves robustness and accuracy, since only facial apperance and lighting variations need to be considered. Even skin toned occlusions such as hands can be handled effectively by our method. We briefly summarize the DDE-based 3D facial regression and then describe how to explicitly incorporate facial segmentation masks.

\paragraph{DDE Regression.} Our facial tracking is performed by regressing a facial shape displacement given the current input RGB image and an initial facial shape from the previous frame. Following the DDE model of~\cite{cao2014displaced}, we represent a facial shape as a linear 3D blendshape model, ($\b_0$, $\B$), with global rigid head motion $(\mathbf{R},\t)$ and 2D residual displacements $\D = [\d_1\ldots\d_m]^T \in \R^{2m}$ of $m=73$ facial landmark positions $\P = [\p_1\ldots\p_m]^T \in \R^{2m}$ (eye contours, mouth, etc.). 
We obtain $\P$ through perspective projection of the 3D face with 2D offsets $\D$:
\begin{equation}\label{equ:displace}
\p_i = \mathrm{\Pi}_f(\mathbf{R}\cdot(\mathbf{b}_0^i + \B^i\mathbf{x})+\t) + \d_i\quad,
\end{equation}
where $\b^i_0$ is the 3D vertex location corresponding to the landmark $\p_i$ in the neutral face $\b_0$, $\B = [\b_1, ..., \b_n]$ the bases of expression blendshapes, $\x \in [0, 1]^n$ the $n = 46$ blendshape coefficients based on FACS~\cite{ekmanfacial}. Each neutral face and expression blendshape is also represented by a linear combination of 50 PCA bases of human identity shapes~\cite{Blanz:1999:MMS} with $[\b_0, \B] = C_r \times \u$, $\u$ the user-specific identity coefficients, and $C_r$ the rank-3 core tensor obtained from the ZJU FaceWarehouse dataset~\cite{cao2014facewarehouse}. We adopt a pinhole camera model, where the projection operator $\mathrm{\Pi}_f: \R^3 \mapsto \R^2$ is specified by a focal length $f$. Thus, we can uniquely determine the 2D landmarks using the shape parameters $\S = \{\mathbf{R}, \t, \x, \D, \u, f\}$. 

While the goal of the regression is to compute all parameters $\S$ given an input frame $\I$, we separate the optimization of the identity coefficients $\u$ and the focal length $f$ from the rest, since they should be invariant over time. Therefore, the DDE regressor only updates the shape vector $\mathbf{Q} = [\mathbf{R}, \t, \x, \D]$ and $[\u,f]$ is computed only in specific key-frames and on a concurrent thread (see~\cite{cao2014displaced} for details). The two-level regressor structure consists of $T$ sequential cascade regressors $\{R_t(\I, \mathbf{Q}_{t})\}_{t=1}^T$ with updates $\delta\mathbf{Q}_{t+1}$ so that $\mathbf{Q}_{t+1} = \mathbf{Q}_t  + \delta \mathbf{Q}_{t+1}$. Each of the weak regressors $R_t$ classifies a set of randomly sampled feature points of $\I$ based on the corresponding pre-trained update vector $\delta\mathbf{Q}_{t+1}$. For each $t$, we sample new sets of 400 feature points via Gaussian distribution on the unit square. Notice that these points are represented as barycentric coordinates of a Delaunay triangulation of the mean of all 2D facial landmarks for improved robustness w.r.t. facial transformations. Each $R_t$ consists of second layer of $K$ primitive cascade regressors based on random ferns of size $F$ (binary decision tree of depth $F$). Each fern regresses a weaker shape parameter update from a feature vector of $F$ pixel intensity differences of feature point pairs from the $400$ samples. The indices of feature point pairs are specified during training by maximizing the correlation to the ground truth regression residuals. The training process also determines the random thresholds and bin classification values of each fern. 

At run-time, as described in~\cite{cao2014displaced}, if a new expression or head pose is observed, we collect the resulting shape parameters $\hat{\S}$ as well as the landmarks $\hat{\P}$, and alternate the updates of the identity coefficients $\u$ and the focal length $f$ by minimizing the offsets $\hat{\D}$ in \refequ{displace} for $L$ collected key-frames until it converges as follows:
\begin{equation}\label{equ:identity_op}
\argmin_{\u, f}{ \sum_{l=1}^L{\sum_{i=1}^m{\Vert \mathrm{\Pi}_f(\mathbf{\hat{R}}_l\cdot(\mathbf{b}_0^i(\u) + \B^i(\u)\cdot\mathbf{\hat{x}}_l)+\hat{\t}_l) - \hat{\p}_{l,i} \Vert^2}}}.
\end{equation}


\paragraph{Training.}


\begin{wrapfigure}{r}{0.45\textwidth}
\vspace{-0.8cm}
\centering
\includegraphics[width=0.43\textwidth]{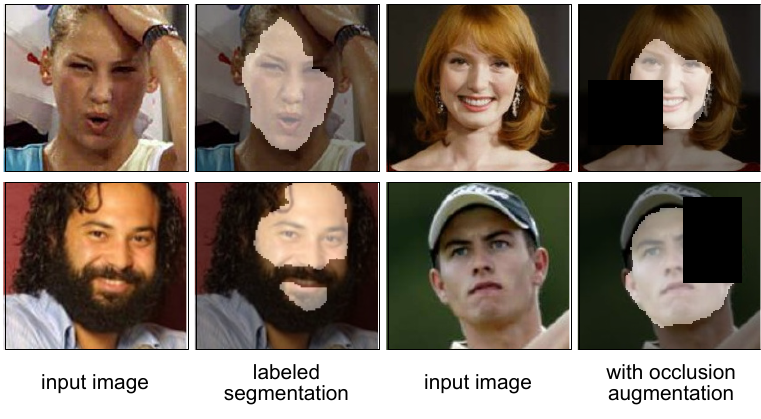}
\caption{Regression training data.
 \label{fig:regtraining}}
\vspace{-0.5cm}
\end{wrapfigure}

The training process consists of constructing the ferns of the primitive regressors and specifying the $F$ pairs of feature point indices based on a large database of facial images with corresponding ground truth facial shape parameters. We construct the ground truth parameters $\{\S_i^g\}_{i=1}^M$ from a set of images $\{\I_i\}_{i=1}^M$ and landmarks $\{\P_i\}_{i=1}^M$. Given landmarks $\P$, the parameters of the ground truth $\S^g$ are computed by minimizing the following objective function $\Theta(\mathrm{R},\t,\x,\u,f)$:
\begin{equation}\label{equ:reconstruction}
\Theta(\mathrm{R},\t,\x,\u,f) = \sum_{i=1}^m{\Vert \mathrm{\Pi}_f(\mathbf{R}\cdot(\mathbf{b}_0^i(\u) + \B^i(\u)\cdot\mathbf{x})+\t) - \p_i \Vert^2}.
\end{equation}

As in~\cite{cao2014displaced}, we use $14,460$ labeled data from FaceWarehouse\cite{cao2014facewarehouse}, LFW\cite{LFWTech}, and GTAV\cite{tarres2012gtav} and learn a mapping from an initial estimation $\S^{*}$ to the ground-truth parameters $\S^g$ given an input frame $\I$. 
An initial set of $N$ shape parameters $\{\S^{*}_i\}_{i=1}^N$ are constructed by perturbing each training parameter in $\S$ within a predefined range. 
Let the suffix $g$ denote the ground-truth value, suffix $r$ a perturbed value. 

We construct the training dataset $\{\S^{*}_i = [\mathbf{Q}_i^r, \u_i^g, f_i^g], \S_i^g = [\mathbf{Q}_i^g, \u_i^g, f_i^g], \I_i\}_{i=1}^N$ and perturb the shape vectors with random rotations, translations, blendshape coefficients as well as, identity coefficients $\u^r$ and the focal length $f^r$ to improved robustness during training. Blendshapes are perturbed $15$ times and the other parameters $5$ times, resulting in a total of $506,100$ training data.
The $T$ cascade regressors $\{R_t(\I, \mathbf{Q}_{t})\}_{t=1}^T$ then update $\mathbf{Q}$ so that the resulting vector $\mathbf{Q}_{t+1} = \mathbf{Q}_t  + \delta \mathbf{Q}_{t+1}$ minimizes the residual to the ground truth $\mathbf{Q}^g$ among all training data $N$. Thus the regressor at stage $t$ is trained as follows:
\begin{equation}\label{equ:fern}
\delta \mathbf{Q}_{t+1} = \argmin_{R}{ \sum_{i = 1}^{N}{ \lVert \mathbf{Q}^{g}_i  - ( \mathbf{Q}_{i,t} + R_t(\I, \mathbf{Q}_{i,t}) ) \rVert_2^2 }}.
\end{equation}

\paragraph{Optimization.} 
For both Equations~\ref{equ:identity_op} and~\ref{equ:reconstruction}, the blendshape and identity coefficients are solved using 3 iterations of non-linear least squares optimization with boundary constraints $\x \in [0, 1]^n$ using an L-BFGS-B solver~\cite{Byrd:1995:LMA} and the rigid motions $(\mathbf{R},\t)$ are obtained by interleaving iterative PnP optimization steps~\cite{Lu:2000:FGC}.

\paragraph{Segmentation-based Regression.}

To incorporate the facial mask $\M$ obtained from~\refsec{probmap} into the regressors $R_t(\I, \P_{t}, \M)$, we simply mark non-face pixels in $\I$ for both training and inference and prevent the regressors to sample features in non-face region. To further enhance the tracking robustness under arbitrary occlusions, which is equivalent to incomplete views after the segmentation process, we augment the training data by randomly cropping out parts on the segmented face images (see \reffig{regtraining}).
For each of the $506,100$ training data sets, we include one additional cropped version with a rectangle centered randomly around the face region with Gaussian distribution and covering up to $80\%$ of the face bounding box in width and height. \reffig{accuracy} and accompanied video shows that this occlusion augmentation significantly improves the robustness under various occlusions after data augmentation.
\vspace{-0.4cm}
\section{Results}\label{sec:result}
\vspace{-0.1cm}
\begin{figure*}[t!]
 \centering
\includegraphics[width=\textwidth]{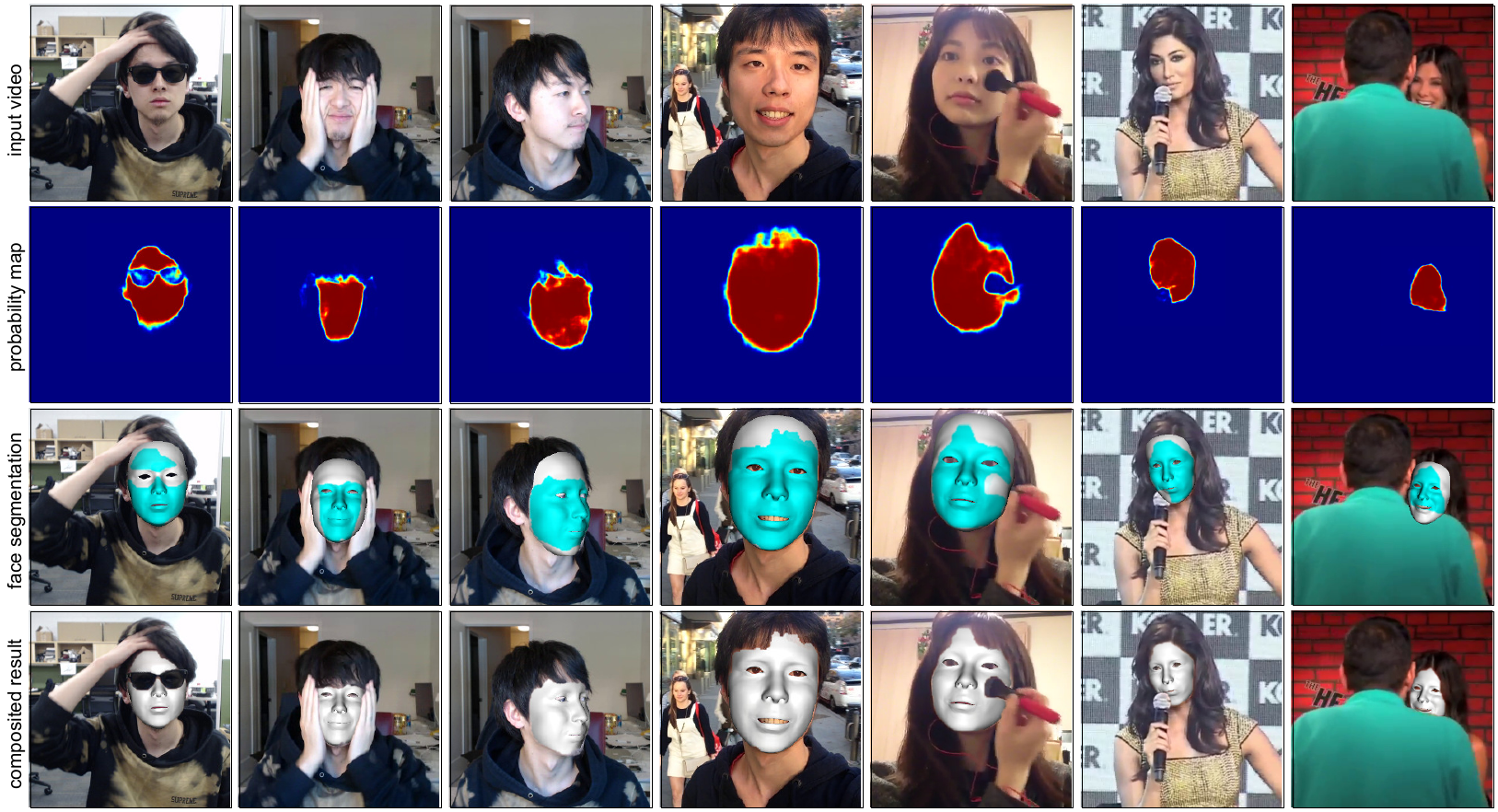}
\caption{ Results. We visualize the input frame, the estimated probability map, the facial segmentation over the tracked template, and the composited result. \vspace{-0.4cm}
 \label{fig:results}}
\end{figure*}

As shown in Figure~\ref{fig:results}, we demonstrate successful facial segmentation and tracking on a wide range of examples with a variety of complex occlusions, including hair, hands, headwear, and props. Our convolutional network effectively predicts a dense probability map revealing face regions even when they are blocked by objects with similar skin tones such as hands. In most cases, the boundaries of the visibile face regions are correctly estimated.  Even when only a small portion of the face is visibile we show that reliable 3D facial fitting is possible when processing input data with clean segmentations. In contrast to most RGB-D based solutions~\cite{hsieh2015unconstrained}, our method works seamlessly in outdoor environments and with any type of video sources.

\paragraph{Segmentation Evaluation and Comparison.}

\begin{wrapfigure}{r}{0.45\textwidth}
\centering
\vspace{-0.85cm}
\includegraphics[width=0.43\textwidth]{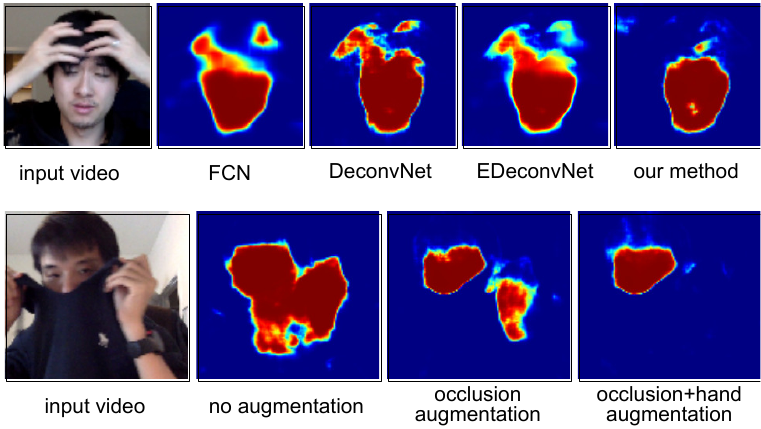}
\caption{Comparison of segmentation result based on different selection of neural network architectures.
 \label{fig:deepeval}}
\vspace{-0.5cm}
\end{wrapfigure}

We evaluate the accuracy of our segmentation technique on 437 color test images from the Caltech Occluded Faces in the Wild (COFW) dataset~\cite{burgos2013robust}.
We use the commonly used intersection over union (IOU) metric between the predicted segmentations and the manually annotated ground truth masks provided by~\cite{jia2014structured}
in order to assess over and under-segmentations. We evaluate our proposed data augmentation strategy as well as the use of negative training samples in Figure~\ref{fig:deepeval} and
show that the explicit use of hand compositings significantly improves the probability map accuracy during hand occlusions. We evalute the architecture of our network in Table~\ref{tab:netcomp} (left) and Figure~\ref{fig:deepeval} and compared our results with the state-of-the-art out of the box segmentation networks, FCN-8s\cite{long_shelhamer_fcn}, DeconvNet~\cite{noh2015learning}, and the naive ensemble of DeconvNet and FCN (EDeconvNet). Compared to FCN-8s and Deconvnet, the IOU of our method is improved by $12.7 \%$ and $1.4 \%$ respectively, but also contains much less noise as shown in Figure~\ref{fig:deepeval}. While comparable to the performance of EDeconvNet, our method achieves nearly double the performance, which enables real-time capabilities (30 fps) on the latest GPU.

We compare in Table~\ref{tab:netcomp} (right), our deep learning-based approach against the current state-of-the-art in facial segmentation: (1) the structured forest technique~\cite{yang2015robust}, (2) the regional predictive power method (RPP)~\cite{jia2014structured} and 3) segmentation-aware part model (SAPM)~\cite{ghiasi2014occlusion,ghiasi2015using}. We measure the IOU and two additional metrics: global (the percentage of all pixels that are correctly classified) and ave(face) (the average recall of face pixels), since the structured forest work~\cite{yang2015robust} uses these two metrics. We demonstrate superior performance to RPP (IOU: 0.833 vs 0.724) and structured forest (global: 0.882 vs 0.839, ave(face): 0.929 vs 0.886), and comparable result to SAPM (IOU: 0.833 vs 0.835, ave(face) 0.929 vs 0.871). Our method is significantly faster than SAPM which requires up to 30s per frame~\cite{ghiasi2014occlusion}.

 \begin{table}[!htb]
    \begin{minipage}[t]{.40\linewidth}
      \vspace{0pt}
      \centering
        \begin{tabular}{ c | c | c }
            Network & mean IOU & FPS\\\hline\hline
            FCN-8s & 0.739 & 37.99\\\hline
            DeconvNet & 0.821 & 44.31\\\hline
            EDeconvNet  & 0.835 & 20.45\\\hline
            Our Method & 0.833 & 43.27\\
        \end{tabular}
        \captionsetup{labelformat=empty,labelsep=none}
    \end{minipage}
    \begin{minipage}[t]{.40\linewidth}
    \vspace{0pt}
      \centering
        \begin{tabular}{ c | c | c | c }
            Method & mean IOU & global & ave(face)\\\hline\hline
            Structured Forest~\cite{jia2014structured} & - & 0.839 & 0.886\\\hline
            RPP~\cite{yang2015robust} & 0.724  & - & - \\\hline
            SAPM~\cite{ghiasi2015using} & 0.835 & 0.886 & 0.871  \\\hline
            Our method & 0.833 & 0.882 & 0.929 \\\hline
            Our Method+GraphCut & 0.839 & 0.887 & 0.927 \\
        \end{tabular}
        \captionsetup{labelformat=empty,labelsep=none}
    \end{minipage} 
    \vspace{0.5cm}
    \caption{segmentation performance for different network structures (left) and state-of-the-art methods (right).}\label{tab:netcomp} \vspace{-0.6cm}
\end{table}

\paragraph{Tracking Evaluation and Comparison.}


\begin{wrapfigure}{r}{0.45\textwidth}
 \centering
\includegraphics[width=0.43\textwidth]{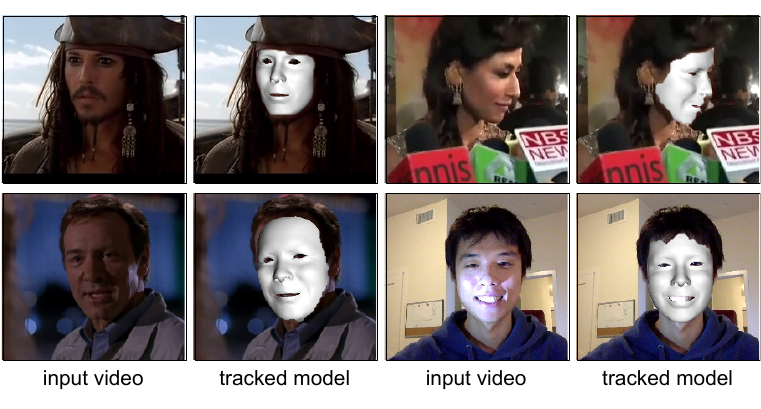}
\caption{Challenging tracking scenes.
 \label{fig:robustness}}
\vspace{0.5cm}
\centering
\includegraphics[width=0.43\textwidth]{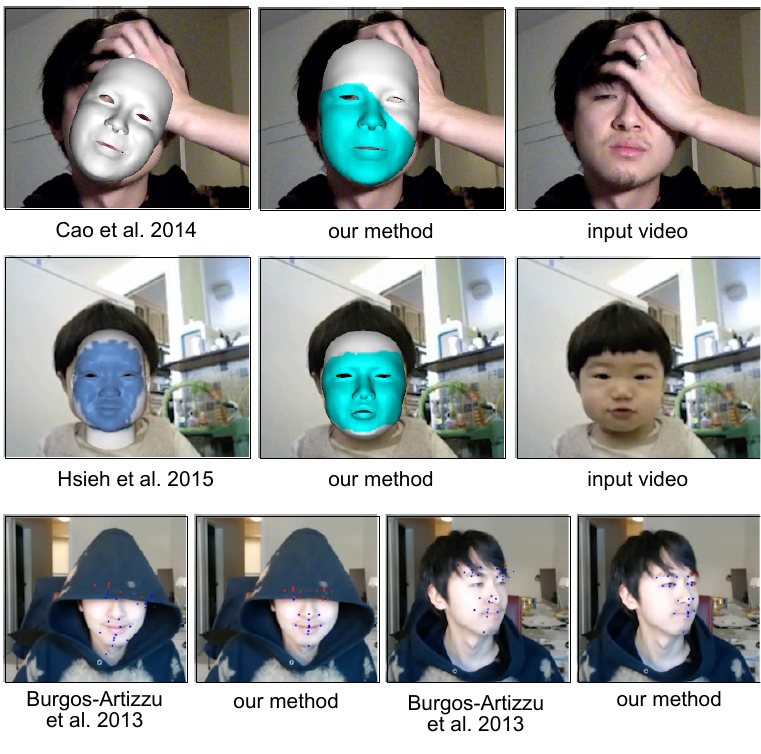}
\caption{Tracking comparison.
 \label{fig:comparison}}
\vspace{-0.5cm}
\end{wrapfigure}

\begin{wrapfigure}{r}{0.5\textwidth}
\vspace{-0.85cm}
 \centering
\includegraphics[width=0.48\textwidth]{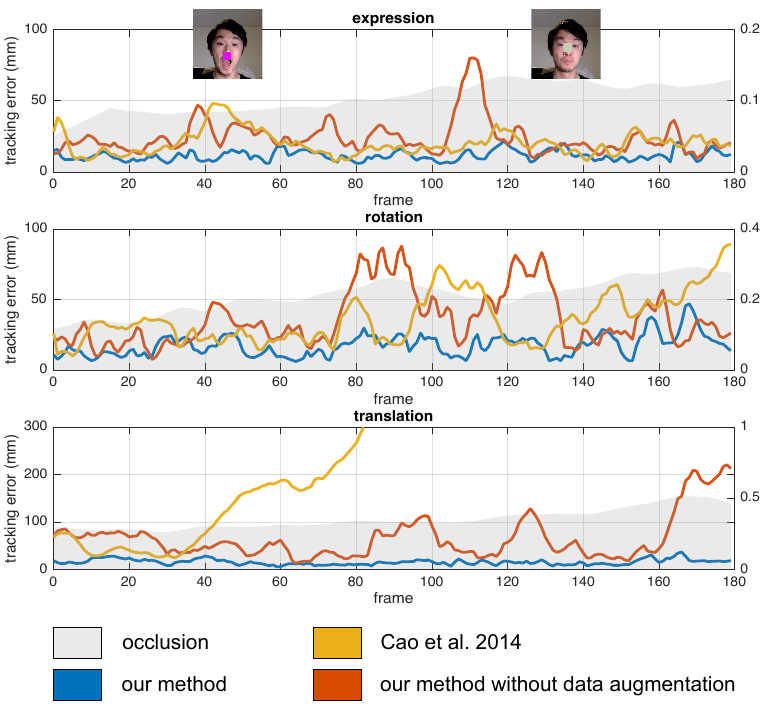}
\caption{Error evaluation on different tracking methods. 
 \label{fig:accuracy}}
\vspace{-0.5cm}
\end{wrapfigure}

In Figure~\ref{fig:robustness}, we highlight the robustness of our approach on extremely challenging cases. Our method can handle difficult lighting conditions, such as shadows and flashlights, as well as side views and facial hair. We further validate our data augmentation strategy during regression training and report quantitative comparisons with the current state-of-the-art method of Cao et al.~\cite{cao2014displaced} in Figure~\ref{fig:accuracy}. Here, we produce an unoccluded face as ground truth and synthetically generated occluding box with increasing size. 
In our experiment, we generated three sequences of 180 frames, covering a wide range of expressions, head rotations and translations. 
We observe that our explicit semantic segmentation approach is critical to ensuring high tracking accuracy. While using the masked training dataset for regression significantly improves robustness, we show that additional performance can be achieved by augmenting this data with additional synthetic occlusions. Figure~\ref{fig:comparison} shows how Cao et al.'s algorithm fails in the presence of large occlusions.  Our method shows comparable occlusion-handling capabilities as the work of~\cite{hsieh2015unconstrained} who rely an RGB-D sensor as input. We demonstrate superior performance to a recent robust 2D landmark estimation method~\cite{burgos2013robust} when comparing the projected landmark positions. 
In particular, our method can handle larger occlusions and head rotations.

\paragraph{Performance.}

Our tracking and segmentation stages run in parallel. The full facial tracking pipeline runs at 30 fps on a quad-core i7 2.8GHz Intel Core i7 with 16GB RAM and the segmentation is offloaded wirelessly to a quad-core i7 3.5GHz Intel Core i7 with 16GB RAM with an NVIDIA GTX Titan X GPU. During tracking, our system takes 18ms to regress the 3D face and 5ms to optimize the identity and the focal length. For segmentation, we measure the following timings: probability map computation 23ms, segmentation refinement 4ms, data transmission 1ms.  run on the GPU, and the remaining implementation is multi-threaded on the CPU.

%
%
%
%

 


\section{Conclusion}\label{sec:conc}

We demonstrate that real-time, accurate pixel-level facial segmentation is possible using only unconstrained RGB images with a deep learning approach. Our experiments confirm that a segmentation network with two-stream deconvolution network and shared convolution network is not only critical for extracting both the overall shape and fine-scale details effectively in real-time,
but also presents the current state-of-the-art in face segmentation. We also found that a carefully designed data augmentation strategy effectively produces sufficiently large training datasets for the CNN to avoid overfitting, especially when only limited ground truth segmentations are available in public datasets. In particular, we demonstrate the first successful facial segmentations for skin-colored occlusions such as hands and arms using composited hand datasets on both positive and negative training samples. Consequently, we show that significantly superior tracking accuracy and robustness to occlusion can be achieved by processing images with masked face regions using a state-of-the-art facial performance capture technique~\cite{cao2014displaced}. Training the DDE regressor with images containing only facial regions and augmenting the dataset with synthetic occlusions ensures continuous tracking in the presence of challenging occlusions (e.g., hair and hands). Although we focus on 3D facial performance capture, we believe the key insight of this paper - reducing the dimensionality using semantic segmentation - is generally applicable to other vision problems beyond facial tracking and regression.

\paragraph{Limitations and Future Work.}

Though surpassing the state-of-the-art, our solution is far from perfect. Since only limited training data is used, the resulting segmentation masks can still yield flickering boundaries. We wish to explore the use of a temporal information, as well as the modeling of domain-specific priors to better handle lighting variations. In addition to facial regions, we would also like to extend our ideas to segment other body parts to facilitate more complex compositing operations that include hands, bodies, and hair.  

%
%
%
%

\section*{Acknowledgements}

We would like to thank Joseph J. Lim, Qixing Huang, Duygu Ceylan, Lingyu Wei, Kyle Olszewski, Harry Shum, and Gary Bradski for the fruitful discussions and the proofreading. We also thank Rui Saito and Frances Chen for being our capture models. This research is supported in part by Adobe, Oculus \& Facebook, Sony, Pelican Imaging, Panasonic, Embodee, Huawei, the Google Faculty Research Award, The Okawa Foundation Research Grant, the Office of Naval Research (ONR) / U.S. Navy, under award number N00014-15-1-2639, the Office of the Director of National Intelligence (ODNI), and Intelligence Advanced Research Projects Activity (IARPA), under contract number 2014-14071600010. The views and conclusions contained herein are those of the authors and should not be interpreted as necessarily representing the official policies or endorsements, either expressed or implied, of ODNI, IARPA, or the U.S. Government. The U.S. Government is authorized to reproduce and distribute reprints for Governmental purpose notwithstanding any copyright annotation thereon.


\bibliographystyle{splncs}
\bibliography{egbib}
\end{document}